\documentclass[11pt,a4paper]{article}
\usepackage[hyperref]{acl2021}
\usepackage{times}
\usepackage{latexsym}

\usepackage{graphicx}
\usepackage{caption}
\usepackage{stfloats}
\usepackage{latexsym}
\usepackage{subcaption}
\usepackage{booktabs}
\usepackage{CJKutf8}
\usepackage{color}
\usepackage{subfiles}
\usepackage{multirow}
\usepackage{amsmath}
\usepackage[ruled]{algorithm2e}
\usepackage{enumitem}
\usepackage{bbold}

\usepackage{microtype}

\aclfinalcopy 
\def\aclpaperid{376} 

\title{DuReader$\rm_\mathbf{robust}$: A Chinese Dataset Towards Evaluating Robustness and Generalization of Machine Reading Comprehension in Real-World Applications}

\author{
Hongxuan Tang\textsuperscript{1}\thanks{\ \ This work was done while the first author was doing internship at Baidu Inc.} \ \ \ 
Hongyu Li\textsuperscript{2} \ \ \  
Jing Liu\textsuperscript{2}\thanks{\ \ Corresponding authors} \ \ \ 
Yu Hong\textsuperscript{1}\textsuperscript{$\dagger$} \ \ \ 
Hua Wu\textsuperscript{2} \ \ \ 
Haifeng Wang\textsuperscript{2}\\
\textsuperscript{1}School of Computer Science and Technology, Soochow University, China\\
\textsuperscript{2}Baidu Inc., Beijing, China\\
 \{{\tt hxtang01, tianxianer\}@gmail.com}\\
 \{{\tt liujing46, lihongyu04, wu\_hua, wanghaifeng\}@Baidu.com}
}

\date{}

\begin{document}
\maketitle

\begin{abstract}

Machine reading comprehension (MRC) is a crucial task in natural language processing and has achieved remarkable advancements. However, most of the neural MRC models are still far from robust and fail to generalize well in real-world applications. In order to comprehensively verify the robustness and generalization of MRC models, we introduce a real-world Chinese dataset -- DuReader$\rm_\mathbf{robust}$. It is designed to evaluate the MRC models from three aspects: over-sensitivity, over-stability and generalization. Comparing to previous work, the instances in DuReader$\rm_\mathbf{robust}$ are natural texts, rather than the altered unnatural texts. It presents the challenges
when applying MRC models to real-world
applications. The experimental results show
that MRC models do not perform well on the challenge test set. Moreover, we analyze the behavior
of existing models on the challenge test
set, which may provide suggestions for future
model development. The dataset and codes are publicly available at \url{https://github.com/baidu/DuReader}.

\end{abstract}

\section{Introduction}

Machine reading comprehension (MRC) requires machines to comprehend text and answer questions about it. With the development of deep learning, the recent studies of MRC have achieved remarkable advancements~\citep{seo2016bidirectional,wang2016machine,devlin2018bert,Liu2019RoBERTaAR,Lan2020ALBERTAL}. However, previous studies show that most of the neural models are not robust enough~\cite{jia2017adversarial,ribeiro2018semantically,wallace2019universal,Welbl2019UndersensitivityIN} and fail to generalize well~\cite{talmor2019multiqa}. 

To further promote the studies of robust and well generalized MRC, we construct a Chinese dataset -- DuReader$\rm_\mathbf{robust}$ which comprises natural questions and documents. In this paper, we focus on evaluating the robustness and generalization from the following aspects, where robustness consists of  over-sensitivity and over-stability: 

\begin{table*}[t]
\small
\begin{subtable}[t]{\textwidth}
    \begin{tabular}{|p{0.475\columnwidth}|p{0.475\columnwidth}|}
    \hline
    \textbf{Passage} & \textbf{Passage} \\
    \begin{CJK*}{UTF8}{gkai}
    近年来，随着琥珀蜜蜡市场的兴起，蜜蜡与琥珀的价格都有不断上涨的趋势，其中蜜蜡首饰的价格一般是琥珀首饰价格的2--4倍，最近几年二者价格差距更大 ……
    \end{CJK*}
    & {\em In recent years, with the rise of the amber market, the price of amber keeps going up. The price of opaque amber is generally 2–4 times the price of clear amber ...} \\
    \hline
    \textbf{Original Question} & \textbf{Original Question} \\
    \begin{CJK*}{UTF8}{gkai}
    琥珀和蜜蜡哪一个比较贵
    \end{CJK*}
    & \em Which is more expensive, clear amber or opaque amber? \\
    \textbf{Golden Answer} :
    \begin{CJK*}{UTF8}{gkai}
    蜜蜡
    \end{CJK*}
    & \textbf{Golden Answer} :  \em opaque amber \\
    \textbf{Predicted Answer} :
    \begin{CJK*}{UTF8}{gkai}
    蜜蜡\  (BERT\textsubscript{base})
    \end{CJK*}
    & \textbf{Predicted Answer} :  {\em opaque amber} (BERT\textsubscript{base}) \\
    \hline
    \textbf{Paraphrase Question} & \textbf{Paraphrase Question} \\
    \begin{CJK*}{UTF8}{gkai}
    蜜蜡和琥珀哪个价格高
    \end{CJK*}
    & \em Which has the higher price, opaque amber or clear amber? \\
    \textbf{Golden Answer} :
    \begin{CJK*}{UTF8}{gkai}
    蜜蜡
    \end{CJK*}
    & \textbf{Golden Answer} :  \em opaque amber \\
    \textbf{Predicted Answer} :
    \begin{CJK*}{UTF8}{gkai}
    琥珀\ (BERT\textsubscript{base})
    \end{CJK*}
    & \textbf{Predicted Answer} :  {\em clear amber} (BERT\textsubscript{base}) \\
    \hline
    \end{tabular}
    \vspace{-2mm}
    \caption[2.0\columnwidth]{An example illustrates the over-sensitivity issue, where BERT\textsubscript{base} gives different predictions to the original question and the paraphrased question.}
    \vspace{1mm}
    \label{oversens-example}
\end{subtable}
\\
\begin{subtable}[t]{\textwidth}
    \begin{tabular}{|p{0.475\columnwidth}|p{0.475\columnwidth}|}
    \hline
    \textbf{Passage} & \textbf{Passage} \\
    \begin{CJK*}{UTF8}{gkai}
    包粽子的线以前人们认为是来自麻叶子，其实是棕榈树，粽子的音就来自棕叶子。
    \end{CJK*}
    & \em Many people argue that the \underline{zongzi (rice dumpling) leaves} \underline{are made of hemp}. Actually, it is the palm tree, the real origin, that endows zongzi with the special pronunciation. \\
    \hline
    \textbf{Question} & \textbf{Question} \\
    \begin{CJK*}{UTF8}{gkai}
    包粽子的线来自什么
    \end{CJK*}
    & \em What is the raw material of zongzi leaves?  \\
    \textbf{Golden Answer} :
    \begin{CJK*}{UTF8}{gkai}
    棕榈树
    \end{CJK*}
    & \textbf{Golden Answer} : \em palm tree \\
    \textbf{Predicted Answer} :
    \begin{CJK*}{UTF8}{gkai}
    麻叶子\ (BERT\textsubscript{base})
    \end{CJK*}
    & \textbf{Predicted Answer} :  {\em hemp} (BERT\textsubscript{base}) \\
    \hline
    \end{tabular}
    \vspace{-2mm}
    \caption{An example illustrates the over-stability issue. The underlined span in the passage appears as a trap because it has many words in common with the question. BERT\textsubscript{base} falls into the trap. }
    \vspace{1mm}
    \label{overstab-example}
\end{subtable}
\\
\begin{subtable}[t]{\textwidth}
    \begin{tabular}{|p{0.475\columnwidth}|p{0.475\columnwidth}|}
    \hline
    \textbf{Passage} & \textbf{Passage} \\
    \begin{CJK*}{UTF8}{gkai}
    {\em cos(2x)'=-sin(2x)*(2x)'=-2sin(2x)}  属于复合函数的求导。
    \end{CJK*}
    & \em cos(2x)'=-sin(2x)*(2x)'=-2sin(2x) This is the derivative of a compound function. \\
    \hline
    \textbf{Question} & \textbf{Question} \\
    \begin{CJK*}{UTF8}{gkai}
    {\em cos2x}的导数是多少?  
    \end{CJK*}
    & \em What is the derivative of cos2x? \\
    \textbf{Golden Answer} :  \em -2sin(2x) & \textbf{Golden Answer} :  \em -2sin(2x) \\
    \textbf{Predicted Answer} :  {\em -sin(2x)} (BERT\textsubscript{base}) & \textbf{Predicted Answer} :  {\em -sin(2x)} (BERT\textsubscript{base}) \\
    \hline
    \end{tabular}
    \vspace{-2mm}
    \caption{An example illustrates the generalization issue. Although BERT\textsubscript{base} is sufficiently trained on large-scale open-domain data, it fails to predict the answer to a math question.}
    \label{generalization-example}
\end{subtable}
\label{intro-example}
\vspace{-2mm}
\caption{The examples of over-sensitivity, over-stability and generalization issues. }
\end{table*}

(1) \textbf{Over-sensitivity} denotes that MRC models provide different
answers to the paraphrased questions. It means that the models are overly sensitive to the difference between the original question and its paraphrased question. We provide an example in Table~\ref{oversens-example}.

(2) \textbf{Over-stability} means that the models might fail into a trap span that has many words in common with the question, and extract an incorrect answer from the trap span. Because the models overly rely on spurious lexical patterns without language understanding. We provide an example in Table~\ref{overstab-example}. 

(3) \textbf{Generalization}. The well-generalized MRC models have good performance on both in-domain and out-of-domain data. Otherwise, they are less generalized. We provide an example in Table~\ref{generalization-example}.

In previous work, the above issues have been studied separately.In this paper, we aim to create a dataset namely DuReader$\rm_\mathbf{robust}$ to comprehensively evaluate the three issues of neural MRC models. 
Previous work mainly studies these issues by altering the questions or the documents. \citet{ribeiro2018semantically,Iyyer2018AdversarialEG,Gan2019ImprovingTR} evaluate the over-sensitivity issue via paraphrase questions generated by rules or generative models. \citet{jia2017adversarial,Ribeiro2018AnchorsHM,Feng2018PathologiesON,wallace2019universal} focus on evaluating the over-stability issue by adding distracting sentences to the documents or reducing question word sequences. However, the altered questions and documents are not natural texts and rarely appear in the real-world applications. It is not clear that how the evaluation based on such unnatural texts can help the improvements of the neural models in real-world applications.
By contrast, all the instances in DuReader$\rm_\mathbf{robust}$ are natural texts and collected from the Baidu search. 

We conduct extensive experiments based on DuReader$\rm_\mathbf{robust}$. The experimental results show that
the models based on pre-trained language models
(LMs) ~\cite{devlin2018bert,sun2019ernie,Liu2019RoBERTaAR} do not perform well on the challenge set. 
Besides, we have the following findings on the behaviors of the models: (1) if a paraphrased question contains more words rephrased from the original question, it is more likely that MRC models provide different answers; (2) the trap spans which share more words with the questions easily mislead MRC models; (3) domain knowledge is a key factor that affects the generalization ability of MRC models.

\section{Dataset: DuReader$\rm_\mathbf{robust}$}

DuReader$\rm_\mathbf{robust}$ is built on DuReader, a large-scale Chinese MRC dataset~\cite{he2017dureader}. 
In DuReader, all questions are issued by real users of Baidu search, and the document-level contexts are collected from search results. In DuReader$\rm_\mathbf{robust}$, we select entity questions and paragraph-level contexts from DuReader. We further employ crowd-workers to annotate the answer span conditioned on the question and the paragraph-level context~\footnote{The instances which have insufficient contexts for answering the questions are discarded.}. Additionally, we used a mechanism to ensure data quality, where 10\% of the annotated data will be randomly selected and reviewed by linguistic experts. If the accuracy is lower than 95\%, the crowd-workers need to revise all the answers until the accuracy for the randomly selected data is higher than 95\%.

\begin{table}[t]
\small
    \centering
    {
    \begin{tabular}{lcccc}
        \hline
        \textbf{Dataset} & \textbf{len(p)} & \textbf{len(q)} & \textbf{len(a)} & \textbf{\#} \\
        \hline
        \textbf{Train} & 291.88 & 9.19 & 5.39 & 14,520 \\
        \textbf{Development} & 288.16 & 9.38 & 6.66 & 1,417 \\
        \textbf{Test} & 285.36 & 9.41 & 6.55 & 1,285 \\
        \textbf{Challenge} & 132.09 & 11.97 & 7.33 & 3,556 \\
        \hline
        \textbf{All} & ~ & ~ & ~ & \textbf{20,778} \\
        \hline
    \end{tabular}
    }
    \caption{The statistics of DuReader$\rm_\mathbf{robust}$.}
    \label{statistics}
\end{table}

Eventually, we collect about 21K instances for DuReader$\rm_\mathbf{robust}$, each of which is a tuple $\langle q, p, A\rangle$, where $q$ is a question, $p$ is a paragraph-level context containing reference answers $A$. 
Similar to the existing MRC datasets, DuReader$\rm_\mathbf{robust}$ consists of training set, in-domain development set and in-domain test set, whose sizes are 15K, 1.4K and 1.3K respectively. Besides, DuReader$\rm_\mathbf{robust}$ contains a challenge test set, in which 3.5K instances are created to evaluate the robustness and generalization of MRC models.
The challenge test set can be divided into three subsets including over-sensitivity set, over-stability set and generalization set.
Table~\ref{statistics} shows the statistics of DuReader$\rm_\mathbf{robust}$.
Besides, DuReader$\rm_\mathbf{robust}$ covers a wide range of answer types (e.g. date, numbers, person, etc. ).
The frequency distribution and examples of the answer types are shown in Table~\ref{ans-type}. 
Next, we will present our way to construct the three subsets in the challenge test set.

\begin{table}[t]
\small
    \centering    
    \begin{tabular}{lcl}
        \hline
        \textbf{Answer Type} & \textbf{\%} & \textbf{Examples} \\
        \hline
        Date & 24.7 &     
        \begin{CJK*}{UTF8}{gbsn}
            15分钟\ (15 minutes)
        \end{CJK*} \\
        Number & 17.5 &
        \begin{CJK*}{UTF8}{gbsn}
            53.28厘米\ (53.28cm)
        \end{CJK*} \\
        Interval & 11.8 &
        \begin{CJK*}{UTF8}{gbsn}
            1\%至5\%\ (1\% to 5\%)
        \end{CJK*} \\
        Person & 8.8 &
        \begin{CJK*}{UTF8}{gbsn}
            成龙\ (Jackie Chan)
        \end{CJK*} \\
        Organization & 7.5 &
        \begin{CJK*}{UTF8}{gbsn}
            湖南卫视\ (Hunan Satellite TV)
        \end{CJK*} \\
        Money & 7.0 &        
        \begin{CJK*}{UTF8}{gbsn}
            2.7亿美元\ (270 million dollars)
        \end{CJK*} \\
        Location & 6.0 &
        \begin{CJK*}{UTF8}{gbsn}
            北京\ (Beijing)
        \end{CJK*} \\
        Software & 2.2 &
        \begin{CJK*}{UTF8}{gbsn}
            百度地图\ (Baidu Map)
        \end{CJK*} \\
        Item & 1.6 &
        \begin{CJK*}{UTF8}{gbsn}
            华为P9\ (Huawei P9)
        \end{CJK*} \\
        Other & 12.9 &
        \begin{CJK*}{UTF8}{gbsn}
            管理学\ (Management Science)
        \end{CJK*} \\
        \hline
    \end{tabular}
    \caption{The frequency distribution and examples of different answer types in DuReader$\rm_\mathbf{robust}$.}
    \label{ans-type}
\end{table}

\subsection{Over-sensitivity Subset}

We build the over-sensitivity subset in the following way. First, we sample a subset of instances $\{\langle q, p, A\rangle\}$ from the in-domain test set of DuReader$\rm_\mathbf{robust}$. For each question $q$, we obtain its $N$ paraphrases $\{q'_{1}, q'_{2}, ..., q'_{N}\}$ using the paraphrase retrieval toolkit (See Appendix~\ref{appendix-kit} for further details). To ensure the paraphrase quality, we employ crowd-workers to discard all false paraphrases. Then, we replace $q$ with the paraphrased question $q'_{i}$, and keep the original context $p$ and answers $A$ unchanged. 
This leads to the new instances $\{\langle q'_{i}, p, A\rangle\}$, and they are used as the model-independent instances in the over-sensitivity subset. Besides, we also employ a model-dependent way to collect instances.
Specifically, we use paraphrased instances to attack the MRC models based on ERNIE~\cite{sun2019ernie} and RoBERTa~\cite{Liu2019RoBERTaAR}. If one of the models gives a different prediction from the predicted answer of the original question, we adopt the instance, otherwise we discard it. 
The instances collected in the above model-dependent and model-independent ways constitute the over-sensitivity subset. 
The over-sensitivity subset consists of 1.2K instances. The number of model-independent instances is equal to that of model-dependent instances. Table~\ref{oversens-example} shows an example in the over-sensitivity subset. 

\begin{algorithm}[t]
\small
\caption{Annotate an instance for over-stability subset}
\label{overstab-algo}
\KwIn{$\{\langle q, p, A\rangle\}$ tuple}
\KwOut{$\{\langle q', p, A'\rangle\}$ tuple or null}
Identify the named entities $\{e_1, ..., e_n\}$ along with their entity types in $p$ \\
Keep the named entities $\{e_i, ..., e_m\}$ with the same types as $A$ \\
\uIf{1 \textless $m$ \textless $k$}{
    \uIf{linguistic experts consider the passage $p$ contains a trap}{
        annotate a new question $q'$ and answers $A'$  \\
        $A$ and $A'$ share the same named entity type \\
        return $\{\langle q', p, A'\rangle\}$
    }
    \lElse{return null}
}
\lElse{return null}
\end{algorithm}

\subsection{Over-stability Subset}

Intuitively, a trap span that has many words in common with the questions may easily mislead MRC models. Following this intuition, the over-stability subset is constructed as follows. First, we randomly select a set of instances $\langle q, p, A\rangle$ from DuReader. In general, a trap span may contain non-answer named entities of the same type as the reference answers $A$. This is because over-stable models usually rely on spurious patterns that  match the correct answer types. Thus, we use a named entity recognizer~\footnote{\url{https://ai.baidu.com/tech/nlp_basic/lexical}} to identify all named entities in $p$ along with their entity types. We keep the corresponding instance, if there are non-answer named entities that are of the same type as $A$. Then, we ask linguistic experts to annotate a new question $q'$ and answers $A'$, if they consider $p$ contains trap spans. $A$ and $A'$ share the same named entity type. The annotated question $q'$ has a high level of lexical overlap with a trap span that does not contain $A$. We say  $\{\langle q', p, A'\rangle\}$ can be considered as a candidate instance. 
Each candidate instance is used to attack one of the MRC models based on ERNIE~\cite{sun2019ernie} and RoBERTa~\cite{Liu2019RoBERTaAR}.
The candidate instance will be used to construct an overstability subset, if one of the model fails.
Algorithm~\ref{overstab-algo} shows the detailed procedure (See Appendix~\ref{appendix-algo} for details). 
As a result, we have 0.8K instances to evaluate over-stability. Table~\ref{overstab-example} shows an  example from the over-stability subset.

\subsection{Generalization Subset}

The in-domain test set consists merely of in-domain data (i.e., the distribution is the same as the one in the  training and development sets).
In order to evaluate the generalization ability of MRC models, we construct a generalization subset which comprises out-of-domain data. 
The out-of-domain data is collected from two vertical domains. The details are as follows.

\textbf{Education} We collect educational questions and documents from Baidu search, and we ask crowdworkers to annotate 1.2K high-quality tuples {$\langle q, p, A\rangle$}. The topics include mathematics, physics, chemistry, language and literature. Table~\ref{generalization-example} shows an example.

\textbf{Finance} Following~\citet{fisch2019mrqa}, we leverage a dataset that was originally designed for information extraction in the finance domain for MRC. We obtain 0.4K instances of financial reports this way. The construction details are presented in Appendix~\ref{financial}.

\section{Experiments}
\label{sec-exp}

\begin{table}[t]
\small
    \centering
    \setlength{\tabcolsep}{3pt}
    {
    \begin{tabular}{lcccccc}
        \hline
        ~ & \multicolumn{2}{c}{\shortstack{\textbf{In-domain} \\ \textbf{dev set}}} &
            \multicolumn{2}{c}{\shortstack{\textbf{In-domain} \\ \textbf{test set}}} &
            \multicolumn{2}{c}{\shortstack{\textbf{Challenge} \\ \textbf{test set}}} \\
        ~ & \textbf{EM} & \textbf{F1} & \textbf{EM} & \textbf{F1} & \textbf{EM} & \textbf{F1} \\
        \hline
        \textbf{BERT\textsubscript{base}} & 71.20 & 82.87 & 67.70 & 80.85 & 37.57 & 53.86 \\
        \textbf{ERNIE 1.0\textsubscript{base}} & 68.73 & 81.12 & 66.72 & 80.50 & 36.75 & 55.64 \\
        \textbf{RoBERTa\textsubscript{large}} & 74.17 & 86.02 & 71.20 & 84.16 & 45.02 & 62.83 \\
        \hline
        \textbf{Human} &  &  & 78.00 & 89.75 & 72.00 & 86.43 \\
        \hline
    \end{tabular}
    }
    \caption{Comparing MRC baselines to human on the development, test and all challenge sets. }
    \label{basic-res}
\end{table}

\begin{table}[t]
\small
    \centering
    \setlength{\tabcolsep}{3pt}
    {
    \begin{tabular}{lcccccc}
        \hline
        ~ & \multicolumn{2}{c}{\shortstack{\textbf{Over-} \\ \textbf{Sensitivity}}} & \multicolumn{2}{c}{\shortstack{\textbf{Over-} \\ \textbf{Stability}}} & \multicolumn{2}{c}{\shortstack{\textbf{Genera-} \\ \textbf{lization}}} \\
        ~ & \textbf{EM} & \textbf{F1} & \textbf{EM} & \textbf{F1} & \textbf{EM} & \textbf{F1} \\
        \hline
        \textbf{BERT\textsubscript{base}} & 53.31 & 69.30 & 16.78 & 38.40 & 36.41 & 50.15 \\
        \textbf{ERNIE 1.0\textsubscript{base}} & 58.10 & 73.89 & 17.27 & 38.34 & 32.86 & 52.84 \\ 
        \textbf{RoBERTa\textsubscript{large}} & 55.24 & 75.16 & 28.18 & 47.03 & 46.03 & 61.67 \\
        \hline
    \end{tabular}
    }
    \caption{The results on the three subsets of the challenge set.}
    \label{robust-res}
\end{table}

\subsection{Baselines and Evaluation Metrics}
We consider three baseline models in the experiments. They are based on different pre-trained language models, including BERT\textsubscript{base}~\cite{devlin2018bert}, ERNIE 1.0\textsubscript{base}~\cite{sun2019ernie}
and RoBERTa\textsubscript{large}~\cite{Liu2019RoBERTaAR}. 
In Appendix~\ref{appendix-hyperpara}, we set the hyperparameters of our  baseline models.

Following~\citet{rajpurkar2016squad}, we use exact match (EM) and F1-score to evaluate the held-out accuracy of an MRC model. 
All the metrics are calculated at Chinese character level, and we normalize both the predicted and true answers by removing spaces and punctuation marks.

\subsection{Main Results}

Table~\ref{basic-res} shows the baseline results on the in-domain development set, in-domain test set, and challenge test set. 
The baseline performance is close to human performance on the in-domain test set, whereas the gap between baseline performance and human performance on the challenge test set is much larger.
In Appendix~\ref{appendix-human-performance}, we describe the method for calculating human performance.

We further evaluate the baselines on the three challenge subsets for over-sensitivity, over-stability and generalization separately. 
Table~\ref{robust-res} shows the results. 
We have found that baseline performance declines significantly for over-stability and generalization subsets (compared to the ``In-domain test set'' in Table~\ref{basic-res}). 
In contrast, the baseline performance degrades less significantly on the over-sensitivity subset, although there is still a noticeable gap.

\subsection{Discussion 1: Over-sensitivity}

First, we calculate the different prediction ratios (DPRs) of the baselines on the over-sensitivity subset.
DPR measures the percentage of the paraphrased questions that yield different predictions.
DPR is formulated in Appendix~\ref{appendix-dpr} . Table~\ref{DPR-res} presents the DPRs of the baselines on the over-sensitivity subset.
The baselines obtained around 16\% to 22\% DPRs, which demonstrates that the baselines are sensitive to part of the paraphrased questions.

\begin{table}[t]
\small
    \centering
    \setlength{\tabcolsep}{10pt}
    {
    \begin{tabular}{lc}
        \hline
        ~ & \textbf{DPR ($\%$)} \\
        \hline
        \textbf{BERT\textsubscript{base}} &  22.73 \\
        \textbf{ERNIE 1.0\textsubscript{base}} & 19.88 \\
        \textbf{RoBERTa\textsubscript{large}} & 16.44 \\
        \hline
    \end{tabular}
    }
    \caption{The DPRs of baselines on the over-sensitivity subset. }
    \label{DPR-res}
\end{table}

\begin{figure}[t]
\centering
\includegraphics[width=0.45\textwidth]{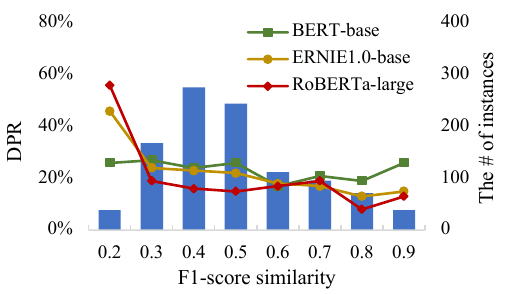}
\caption{The correlation between DPR and F1-score based question similarity on the over-sensitivity subset.}
\label{oversens-exp}
\end{figure}

Second, we examine a hypothesis - if a paraphrased question contains more words rephrased from the original question, the MRC model is more likely to produce different answers.
To measure how similar paraphrased questions are to the original questions, we use the F1-score. 
A low F1-score means that many words in the original  question have been rephrased. 
We divide the paraphrased questions into buckets based on how similar they are to the original questions, and we then examine whether there is correlation between DPR and F1-score similarity. 
Based on Figure~\ref{oversens-exp}, we can observe that the DPRs of all the baselines are negatively correlated with the F1-score similarity between the original and  paraphrased questions. The results confirm the hypothesis.

\subsection{Discussion 2: Over-stability}

\begin{figure}[t]
\centering
\includegraphics[width=0.43\textwidth]{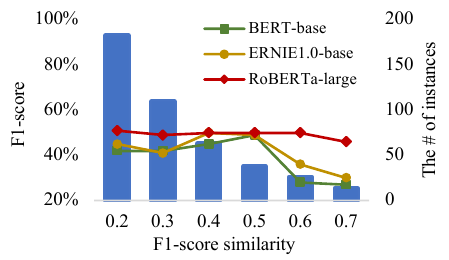}
\caption{The correlation between the model performance and question-trap similarity on the over-stability subset.}
\label{overstab-exp}
\end{figure}

MRC models might be easily misled by trap spans that share many words with the questions. 
We examine whether there is a correlation between MRC performance (F1-score) and question-trap similarity in this section. 
Based on the similarity between trap spans and questions, we divide trap spans into buckets.
According to Figure~\ref{overstab-exp}, the performance of the base models decreases as similarity increases  and the large model (RoBERTa\textsubscript{large}) is less over-stable than the base ones.

\subsection{Discussion 3: Generalization}
Table~\ref{general-res} shows the baseline performance in the the domains of finance and education. We can observe that the baselines perform poorly for both domains. 
Additionally, we examine how baseline models behave in the education domain. 
Table~\ref{res-edu} shows the performance of RoBERTa\textsubscript{large} on the four topics in the education domain. 
The model performs much worse when it comes to math and chemistry, since these topics are rare in the training set. 
The results of this analysis suggest that domain knowledge is a key factor affecting the generalization ability of MRC models. 
More discussion can be found in Appendix~\ref{appendix-ana}.

\section{Conclusion}

In this paper, we create a Chinese dataset -- DuReader$\rm_\mathbf{robust}$ and use it to evaluate both the robustness and generalization of the MRC models. Its questions and documents are natural texts from Baidu search. This presents the robustness and generalization challenges in the real-world applications. Our experiments show that the MRC models based on the pre-trained LMs do not perform well on DuReader$\rm_\mathbf{robust}$ challenge set. We also conduct extensive experiments to examine the behaviors of the MRC models on the dataset and provide insights for future model development.

\begin{table}[t]
\small
    \centering
    {
    \begin{tabular}{lcccc}
        \hline
        ~ & \multicolumn{2}{c}{\textbf{Finance}} & \multicolumn{2}{c}{\textbf{Education}} \\
        ~ & \textbf{EM} & \textbf{F1} & \textbf{EM} & \textbf{F1} \\
        \hline
        \textbf{BERT\textsubscript{base}} & 30.73 & 51.16 & 38.70 & 50.83 \\
        \textbf{ERNIE 1.0\textsubscript{base}} & 26.53 & 50.53 & 34.67 & 53.11  \\
        \textbf{RoBERTa\textsubscript{large}} & 40.22 & 61.16 & 47.77 & 61.82 \\
        \hline
    \end{tabular}
    }
    \caption{The performance of baselines in the domains of education and finance.}
    \label{general-res}
\end{table}

\begin{table}[t]
\small
    \centering
    \begin{tabular}{lccc}
         \hline
         \textbf{Topcis} & \textbf{EM} & \textbf{F1} & \textbf{\#} \\
         \hline
         \textbf{Math} & 19.85 & 34.63 & 136 \\
         \textbf{Chemistry} & 37.46 & 53.88 & 323 \\
         \textbf{Language} & 44.31 & 61.18 & 255 \\
         \textbf{Others} & 69.63 & 79.28 & 438 \\
         \hline
         \textbf{All} & 49.13 & 62.88 & 1152 \\
         \hline
    \end{tabular}
    \caption{The performance of baselines on different topics in the domain of education.}
    \label{res-edu}
\end{table}

\section*{Acknowledgement}

This research work is supported by the National Key Research and Development Project of China (No. 2018AAA0101900) and National Natural Science Foundation of China (Grants No.61773276 and No.62076174). The authors would like to thank the anonymous reviewers for their insightful comments and suggestions. Jing Liu and Yu Hong are the corresponding authors of the paper.

\section*{Ethical Considerations}

We aim to provide researchers and developers with a dataset DuReader$\rm_\mathbf{robust}$ to improve the robustness and generalization ability of MRC models.
We also take the potential ethical issues into account. (1) All the instances in the DuReader$\rm_\mathbf{robust}$ have been desensitised. (2) Regarding to the issue of labor compensation, we make sure that all the crowdsourcing workers are fairly compensated.

\bibliographystyle{acl_natbib}
\bibliography{anthology,acl2021}

\clearpage
\appendix

\section{Paraphrase Retrieval Toolkit}
\label{appendix-kit}

We use a paraphrase retrieval toolkit to obtain paraphrase questions. 
The toolkit is used internally at Baidu, and our manual evaluations show that the  accuracy of the retrieval results is around 98\%. 
The paraphrase retrieval toolkit consists of two basic modules as follows.

\begin{itemize}[leftmargin=*]
\item {\textbf{Paraphrase Candidate Retriever} The retriever is a light-weight module. Given a question, the retriever will retrieve top-k paraphrase candidates from the search logs of Baidu Search. }
\item {\textbf{Paraphrase Candidate Re-ranker} The re-ranker is a model fine-tuned from ERNIE~\cite{sun2019ernie}  by using a set of manually labeled paraphrase questions. Given a set of retrieved paraphrase candidates, the re-ranker will estimate the semantic similarity between the original question and the paraphrase candidates. If the semantic similarity is higher than a pre-defined threshold, the candidate will be used as a paraphrased question. }
\end{itemize}

\section{The Illustration of Annotating Over-stability Instances}
\label{appendix-algo}

\begin{figure}[th]
\centering
\includegraphics[width=0.47\textwidth]{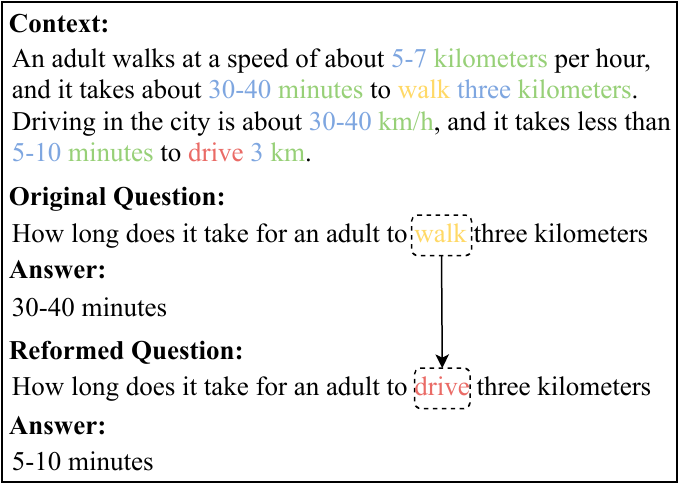}
\caption{The illustration of annotating an over-stability instance. 
}
\label{overstab-anno}
\end{figure}

Figure~\ref{overstab-anno} illustrates the annotation of an over-stability instance. In the instance, the answer to the original question is 30-40 minutes. The entity type of 5-10 minutes is the same as 30-40 minutes. The annotator raise a new question by revising the original question, the answer to the new question is 5-10 minutes. The sentence contains 30-40 minutes has many words in common with the new question, and it is considered as a trap sentence. The new question may mislead the model to predict the answer to the new question as 30-40 minutes. 

\section{The Construction of Finance Data}
\label{financial}
We leverage a dataset that is originally designed for information extraction in finance domain. The original dataset contains the full texts of the financial reports as documents and the structured data that is extracted from the texts. Then, we use templates to generate questions for each data field in the structured data. Finally, we use these constructed instances for MRC. Each instance contains (1) a question generated from a template for a data field, (2) an answer that is the value in the data field and (3) a document from which the value (i.e. answer) is extracted.

\section{Hyperparameters}
\label{appendix-hyperpara}
We use a number of pre-trained language models in our baseline systems. When fine-tuning different pre-trained language models, we use the same hyperparameters. The settings of hyperparameters are as follows. The learning rate is set to 3e-5 and the batch size is $32$. We set the number of epochs to $5$. The maximal answer length and document length are  set to $20$ and $512$, respectively. We set the length of document stride to $128$. All experiments are conducted on 4 Tesla P40 GPUs. 

\section{Human Performance}
\label{appendix-human-performance}
We evaluate human performance on both the in-domain test set and challenge test set. We randomly sample two hundred instances from the in-domain test set, and three hundred instances from the challenge test set. 
We ask crowdworkers to provide answers to the questions in the sampled instances. Then, we use EM and F1-scores of these annotated answers as human performance. 

\section{Different Prediction Rate}
\label{appendix-dpr}
Different prediction rate (DPR) measures the percentage of paraphrase questions whose predictions are different from the original questions. Formally, we define DPR of a neural model $f(\theta)$ on a dataset $D$ as follows. 

\begin{equation*}
DPR_{D}(f(\theta)) = \frac{\sum\limits_{(q, q\prime) \in Q} \mathbb{1}[f(\theta; q) \neq f(\theta; q\prime)]}{\| Q \|}
\end{equation*}
where, $f(\theta;q)$ denotes the prediction of the trained MRC model $f(\theta)$. $Q$ represents a set of pairs of original question $q$ and paraphrased question $q\prime$ in dataset $D$, and $\mathbb{1}[*]$ is an indicator function. A high DPR score means that the MRC model is overly sensitive to the paraphrased question $q\prime$, otherwise insensitive.

\section{Experimental Analysis}
\label{appendix-ana}

\subsection{Over-sensitivity Analysis}
\label{ana-oversens}

We further analyze the prediction results to figure out what kind of paraphrases lead to different predictions. Five types of paraphrasing phenomena have been found, including (1) word reordering (WR), (2) replacement of function words (RF), (3) substitution by synonyms (SS), (4) inserting or removing content words (AD), and (5) more than one previously defined types happen in one paraphrase (CO). 
We randomly sample one hundred instances from the over-sensitivity subset and analyze the changes of the predictions by ERNIE~\cite{sun2019ernie}. 
As shown in Table~\ref{oversens_anno}, most of changed predictions come from AD and CO. This analysis suggests that the models are sensitive to the changes of content words. 

\begin{table}[t]
\small
    \centering
    {
    \begin{tabular}{lcc}
        \hline
        \textbf{Types} & \textbf{\# of changes (\%)} & \textbf{\# of same (\%)}
        \\
        
        \hline
        \multirow{1}{*}{\textbf{WR}} &
        \multirow{1}{*}{1 (12.50)} &
        \multirow{1}{*}{7 (87.50)} 
        \\
        
        \multirow{1}{*}{\textbf{RF}} &
        \multirow{1}{*}{0 (00.00)} &
        \multirow{1}{*}{4 (100.00)}
        \\
        
        \multirow{1}{*}{\textbf{SS}} &
        \multirow{1}{*}{6 (17.14)} &
        \multirow{1}{*}{29 (82.85)}
        \\

        \multirow{1}{*}{\textbf{AD}} &
        \multirow{1}{*}{7 (23.33)} &
        \multirow{1}{*}{23 (76.66)}
        \\
        
        \multirow{1}{*}{\textbf{CO}} &
        \multirow{1}{*}{7 (30.43)} &
        \multirow{1}{*}{16 (69.56)}
        \\
        \hline
    \end{tabular}
    }
    \caption{Distributions of paraphrases and DPRs.}
    \label{oversens_anno}
\end{table}

\begin{table}[tbp]
\small
    \centering
    \setlength{\tabcolsep}{5.5pt}
    {
    \begin{tabular}{lccc}
         \hline
         \textbf{Question Types} & \textbf{EM} & \textbf{F1} & \textbf{\#} \\
         \hline
         \textbf{Company abbreviations} & 0 & 31.15 & 18 \\
         \textbf{Pledgee} & 80.76 & 89.96 & 26  \\
         \textbf{Pledgor} & 0 & 24.62 & 25 \\
         \textbf{The pledge amount} & 18.36 & 53.84 & 98 \\
         \textbf{Others (e.g. pledge date)} & 47.91 & 58.97 & 48 \\
         \hline
         \textbf{All} & 28.83 & 54.05 & 215 \\
         \hline
    \end{tabular}
    }
    \caption{The performance of RoBERTa\textsubscript{large} on the five topics in the domain of financial reports. }
    \label{res-ple}
\end{table}

\subsection{Generalization Analysis}
\label{ana-gene}

In previous section, we have already analyzed the behaviors of baseline systems on education domain. In this section, we conduct analysis on financial domain. The data of financial domain contains management changes and equity pledge. The performance of RoBERTa\textsubscript{large} on management changes and equity pledge is 68.63\% and 49.15\% respectively. The model generalizes well on management changes, since the training set contains the relevant knowledge about asking person names. By contrast, the model performs worse on equity pledge. We classify the instances of equity pledge into five sets according to the question types. Table~\ref{res-ple} shows the performance of RoBERTa\textsubscript{large} on the five question types. We can observe that the model performs the worst on the questions about company abbreviations, pledgee and pledgor, since there is little domain knowledge in the training set. By contrast, the model performs better on the questions about amount and date, since the model has already learnt relevant knowledge in the training set.

\end{document}